\documentclass[11pt,a4paper]{article}
\usepackage[hyperref]{emnlp2018}
\usepackage{times}
\usepackage{latexsym}

\usepackage{microtype}
\usepackage{graphicx}
\usepackage{subfigure}
\usepackage{amsmath,amsfonts}       
\usepackage{booktabs} 
\usepackage{xspace}
\usepackage{url}
\usepackage{multirow}
\usepackage[title]{appendix}

 \aclfinalcopy

\DeclareMathOperator*{\argmin}{arg\,min}

\def \ie {\emph{i.e.}\xspace}
\def \etal {\emph{et al.}\xspace}

\title{Loss in Translation: \\ Learning Bilingual Word Mapping with a Retrieval Criterion}

\author{Armand Joulin \\
  \And
  Piotr Bojanowski \\
  \And
  Tomas Mikolov \\
  Facebook AI Research \\
  {\tt \{ajoulin,bojanowski,tmikolov,rvj,egrave\}@fb.com} \\
  \And
  Herv\'e J\'egou \\
  \And
  Edouard Grave \\
  }

\date{}

\begin{document}
\maketitle
\begin{abstract}
Continuous word representations learned separately on distinct languages can be aligned
so that their words become comparable in a common space.
Existing works typically solve a least-square regression problem to learn a rotation aligning a small bilingual lexicon,
and use a retrieval criterion for inference.
In this paper, we propose an unified formulation that directly optimizes a retrieval criterion in an end-to-end fashion.
Our experiments on standard benchmarks show that our approach outperforms the state of the art on word translation, with the biggest improvements observed for distant language pairs such as English-Chinese.
\end{abstract}

\section{Introduction}
\label{sec:intro}

Previous work has proposed to learn a linear mapping between continuous representations of words by employing a small bilingual lexicon as supervision. 
The transformation generalizes well to words that are not observed during training, making it possible to extend the lexicon. Another application is to transfer predictive models between languages~\cite{klementiev2012inducing}.  

The first simple method proposed by~Mikolov \etal \shortcite{mikolov2013exploiting} has been subsequently improved by changing the problem parametrization.
One successful suggestion is to $\ell_2$--normalize the word vectors and to constrain the linear mapping to be orthogonal~\cite{xing2015normalized}.
An alignment is then efficiently found using orthogonal Procrustes~\cite{artetxe2016learning,smith2017offline}, 
improving the accuracy on standard benchmarks.

Yet, the resulting models suffer from the so-called ``hubness problem'': some word vectors tend to be the nearest neighbors of an abnormally high number of other words.
This limitation is now addressed 
by applying a corrective metric at inference time, such as the inverted softmax (ISF)~\cite{smith2017offline} or the cross-domain similarity local scaling (CSLS)~\cite{conneau2017word}.
This is not fully satisfactory because the loss used for inference is not consistent with that employed for training. 
This observation suggests that the square loss is suboptimal and could advantageously be replaced by a loss adapted to retrieval. 

In this paper, we propose a training objective inspired by the CSLS retrieval criterion. 
We introduce convex relaxations of the corresponding objective function, which are efficiently optimized with projected subgradient descent.
This loss can advantageously include unsupervised information and therefore leverage the representations of words  not occurring in the training lexicon. 

Our contributions are as follows. 
First we introduce our approach and empirically evaluate it on standard benchmarks for word translation. We obtain state-of-the-art bilingual mappings for more than $25$ language pairs.
Second, we specifically show the benefit of our alternative loss function and of leveraging  unsupervised information. 
Finally, we show that with our end-to-end formulation, a non-orthogonal mapping achieves better results. 
The code for our approach is a part of the fastText library\footnote{\url{https://github.com/facebookresearch/fastText/tree/master/alignment/}} and the aligned vectors are available on \url{https://fasttext.cc/}.

\section{Preliminaries on bilingual mappings}
\label{sec:prelim}

This section introduces pre-requisites and prior works to learn a mapping between two languages, using a small bilingual lexicon as supervision.

We start from two sets of continuous representations in two languages, each learned on monolingual data.
Let us introduce some notation.
Each word $i \in \{1, \dots, N\}$ in the source language (respectively target language) is associated with a vector $\mathbf{x}_i \in \mathbb{R}^d$
(respectively $\mathbf{y}_i \in \mathbb{R}^d$).
For simplicity, we assume that our initial lexicon, or seeds, corresponds to the first $n$ pairs $(\mathbf{x}_i, \mathbf{y}_i)_{i \in \{1, \dots, n\}}$.
The goal is to extend the lexicon to all source words~$i \in \{n+1, \dots, N\}$ that are not seeds.
Mikolov \etal \shortcite{mikolov2013exploiting} learn a linear mapping~$\mathbf{W} \in \mathbb{R}^{d \times d}$ between the word vectors of the seed lexicon that minimizes a measure of discrepancy between mapped word vectors of the source language and word vectors of the target language:
\begin{equation}
  \min_{\mathbf{W}\in\mathbb{R}^{d\times d}} \quad \frac{1}{n} \sum_{i=1}^{n} \ell(\mathbf{W x}_i, \mathbf{y}_i),
  \label{eq:emb_align}
\end{equation}
where $\ell$ is a loss function, typically the square loss $\ell_2(\mathbf{x}, \mathbf{y})=\|\mathbf{x}-\mathbf{y}\|_2^2$.
This leads to a least squares problem, which is solved in closed form.

\paragraph{Orthogonality.}
The linear mapping $\mathbf{W}$ is constrained to be orthogonal, \ie such that $\mathbf{W}^\top \mathbf{W} = \mathbf{I}_d$,
where $\mathbf{I}_d$ is the $d$-dimensional identity matrix.
This choice preserves distances between word vectors, and likewise word similarities.
Previous works~\cite{xing2015normalized,artetxe2016learning,smith2017offline}
experimentally observed that constraining the mapping in such a way improves the quality of the inferred lexicon.
With the square loss and by enforcing an orthogonal mapping $\mathbf{W}$, Eq.~(\ref{eq:emb_align}) admits a
closed form solution~\cite{gower2004procrustes}:
$
\mathbf{W}^* = \mathbf{UV}^\top,
$
where $\mathbf{UDV}^{\top}$ is the singular value decomposition of the matrix $\mathbf{Y}^{\top}\mathbf{X}$.

\paragraph{Inference.} Once a mapping $\mathbf{W}$ is learned,
one can infer word correspondences for words that are not in the initial lexicon.
The translation $t(i)$ of a source word $i$ is obtained as
\begin{equation}
  t(i) \in \argmin_{j \in \{1, \dots, N\}} \ \ell(\mathbf{W x}_i, \mathbf{y}_j).
\end{equation}
When the squared loss is used, this amounts to computing $\mathbf{Wx}_i$ and to performing a nearest neighbor search with respect  to the Euclidean distance:
\begin{equation}
  t(i) \in \argmin_{j \in \{1, \dots, N\}} \| \mathbf{Wx}_i - \mathbf{y}_j \|_2^2.
  \label{eq:infer}
\end{equation}

\paragraph{Hubness.}
A common observation is that nearest neighbor search for bilingual lexicon inference suffers from the ``hubness problem''~\cite{doddington1998sheep,dinu2014improving}.
Hubs are words that appear too frequently in the neighborhoods of other words.
To mitigate this effect, a simple solution is to replace, at inference time, the square $\ell_2$-norm in Eq.~(\ref{eq:infer}) by another criterion, such as ISF~\cite{smith2017offline} or CSLS~\cite{conneau2017word}.

This solution, both with ISF and CSLS criteria, is applied with a transformation $\mathbf{W}$ learned using the square loss.
However, replacing the loss in Eq.~(\ref{eq:infer}) creates a discrepancy between the learning of the translation model and the inference.

\section{Word translation as a retrieval task}
\label{sec:approach}

In this section, we propose to directly include the CSLS criterion in the model in order to make learning and inference consistent.
We also show how to incorporate unsupervised information..

The CSLS criterion is a similarity measure between the vectors~$\mathbf{x}$ and $\mathbf{y}$ defined as:
\begin{multline*}
  \textsc{csls}(\mathbf{x}, \mathbf{y}) = - 2 \cos(\mathbf{x}, \mathbf{y}) \\
  + \frac{1}{k} \sum_{\mathbf{y}' \in \mathcal{N}_Y (\mathbf{x})} \cos(\mathbf{x}, \mathbf{y}')
  + \frac{1}{k} \sum_{\mathbf{x}' \in \mathcal{N}_X (\mathbf{y})} \cos(\mathbf{x}', \mathbf{y}),
\end{multline*}
where $\mathcal{N}_Y (\mathbf{x})$ is the set of $k$ nearest neighbors of the point $\mathbf{x}$ in the set of target word vectors~${Y=\{\mathbf{y}_1, \dots, \mathbf{y}_N\}}$,
and $\cos$ is the cosine similarity.
Note, the second term in the expression of the CSLS loss does not change the neighbors of $\mathbf{x}$.
However, it gives a loss function that is symmetrical with respect to its two arguments, which is a desirable property for training.

\paragraph{Objective function.} Let us now write the optimization problem for learning the bilingual mapping with CSLS.
At this stage, we follow previous work and constrain the linear mapping $\mathbf{W}$ to belong to the set of orthogonal matrices $\mathcal{O}_d$.
Here, we also assume that word vectors are $\ell_2$-normalized.
Under these assumptions, we have $\cos(\mathbf{W x}_i, \mathbf{y}_i) = \mathbf{x}_i^\top \mathbf{W}^\top \mathbf{y}_i$. Similarly, we have
$
\| \mathbf{y}_j - \mathbf{Wx}_i \|_2^2 = 2 - 2 \mathbf{x}_i^\top \mathbf{W}^\top \mathbf{y}_j.
$
Therefore, finding the $k$ nearest neighbors of $\mathbf{Wx}_i$ among the elements of $Y$ is equivalent to finding the $k$ elements of $Y$ which have the largest dot product with $\mathbf{Wx}_i$.
We adopt this equivalent formulation because it leads to a convex formulation when relaxing the orthogonality constraint on $\mathbf{W}$.
In summary, our optimization problem with the Relaxed CSLS loss (RCSLS) is written as:
\begin{align}
  \min_{\mathbf{W} \in \mathcal{O}_d} & \frac{1}{n} \sum_{i = 1}^n - 2 \mathbf{x}_i^\top \mathbf{W}^\top \mathbf{y}_i \nonumber \\
  + & \frac{1}{k} \sum_{\mathbf{y}_j \in \mathcal{N}_Y (\mathbf{W x}_i)} \mathbf{x}_i^\top \mathbf{W}^\top \mathbf{y}_j \nonumber \\
  + & \frac{1}{k} \sum_{\mathbf{W} \mathbf{x}_j \in \mathcal{N}_X (\mathbf{y}_i)} \mathbf{x}_j^\top \mathbf{W}^\top \mathbf{y}_i.
  \label{eq:prob}
\end{align}

\paragraph{Convex relaxation.}
\label{sec:optim}
Eq.~(\ref{eq:prob}) involves the minimization of a non-smooth cost function over the manifold of orthogonal matrices $\mathcal{O}_d$.
As such, it can be solved using manifold optimization tools~\citep{manopt}.
In this work, we consider as an alternative to the set $\mathcal{O}_d$, its convex hull $\mathcal{C}_d$, \ie, the unit ball of the spectral norm.
We refer to this projection as the ``Spectral'' model.
We also consider the case where these constraints on the alignment matrix are simply removed.

Having a convex domain allows us to reason about the convexity of the cost function.
We observe that the second and third terms in the CSLS loss can be rewritten as follows:
\begin{equation}\label{eq:max}
  \sum_{\mathbf{y}_j \in \mathcal{N}_k (\mathbf{W x}_i)} \mathbf{x}_i^\top \mathbf{W}^\top \mathbf{y}_j =
  \max_{S \in \mathcal{S}_k(n)} \sum_{j \in S} \mathbf{x}_i^\top \mathbf{W}^\top \mathbf{y}_j \nonumber,
\end{equation}
where $\mathcal{S}_k(n)$ denotes the set of all subsets of $\{1, \dots, n\}$ of size $k$.
This term, seen as a function of $\mathbf{W}$, is a maximum of linear functions of $\mathbf{W}$, which is convex~\citep{boyd2004convex}.
This shows that our objective function is convex with respect to the mapping $\mathbf{W}$ and piecewise linear (hence non-smooth).
Note, our approach could be generalized to other loss functions by replacing the term $\mathbf{x}_i^\top \mathbf{W}^\top \mathbf{y}_j$ by any function convex in $\mathbf{W}$.
We minimize this objective function over the convex set $\mathcal{C}_d$ by using the projected subgradient descent algorithm.

The projection onto the set $\mathcal{C}_d$ is solved by taking the singular value decomposition (SVD) of the matrix, and thresholding the singular values to one.

\paragraph{Extended Normalization.}
\label{sec:refinement}
Usually, the number of word pairs in the seed lexicon $n$ is small with respect to the size of the dictionaries $N$.
To benefit from unlabeled data, it is common to add an iterative ``refinement procedure''~\citep{artetxe2017learning} when learning the translation model $\mathbf{W}$.
Given a model $\mathbf{W}_t$, this procedure iterates over two steps. First it augments the training lexicon by keeping the best-inferred translation in Eq.~(\ref{eq:infer}). Second it learns a new mapping $\mathbf{W}_{t+1}$ by solving the problem in Eq.~(\ref{eq:emb_align}).
This strategy is similar to standard semi-supervised approaches where the training set is augmented over time.
In this work, we propose to use the unpaired words in the dictionaries as ``negatives'' in the RCSLS loss:
instead of computing the $k$-nearest neighbors $\mathcal{N}_Y(\mathbf{Wx}_i)$ amongst the annotated words $\{\mathbf{y}_1, \dots, \mathbf{y}_n\}$, we do it over the whole dictionary $\{\mathbf{y}_1, \dots, \mathbf{y}_N\}$.


\begin{table*}[t]
\setlength{\tabcolsep}{4.8pt} 
\centering
\begin{tabular}{l cc cc cc cc  cc c}
\toprule
Method              &en-es&es-en&en-fr&fr-en&en-de&de-en&en-ru&ru-en&en-zh&zh-en& avg.\\
\midrule
Adversarial + refine  & 81.7& 83.3& 82.3& 82.1& 74.0& 72.2& 44.0& 59.1& 32.5 & 31.4 & 64.3\\
ICP + refine          & 82.2& 83.8& 82.5& 82.5& 74.8& 73.1& 46.3& 61.6&  - & - & - \\
Wass. Proc. + refine  & 82.8& 84.1& 82.6&82.9&75.4&73.3&43.7&59.1& - & - & - \\
\midrule
Least Square Error  & 78.9 & 80.7 & 79.3 & 80.7 & 71.5 & 70.1 & 47.2 & 60.2 & 42.3 & 4.0 & 61.5\\
Procrustes          & 81.4 & 82.9 & 81.1 & 82.4 & 73.5 & 72.4 & 51.7 & 63.7 &42.7 & 36.7& 66.8\\
Procrustes + refine & 82.4 & 83.9 & 82.3 & 83.2 & 75.3 & 73.2 & 50.1& 63.5 & 40.3& 35.5 & 66.9 \\
\midrule
RCSLS + spectral  & 83.5 & 85.7 & 82.3  & \textbf{84.1} & 78.2 & 75.8  & 56.1  & 66.5 & 44.9  & 45.7 & 70.2\\
RCSLS             & \textbf{84.1} & \textbf{86.3} &  \textbf{83.3} & \textbf{84.1} &  \textbf{79.1} & \textbf{76.3}  & \textbf{57.9}  &  \textbf{67.2} & \textbf{45.9}  & \textbf{46.4} & \textbf{71.0}\\
\bottomrule
\end{tabular}
\caption{
  Comparison between RCSLS, Least Square Error, Procrustes and unsupervised approaches in the setting of~\citet{conneau2017word}.
  All the methods use the CSLS criterion for retrieval.
  ``Refine'' is the refinement step of~\citet{conneau2017word}.
  Adversarial, ICP and Wassertsein Proc. are unsupervised~\citep{conneau2017word, hoshen2018iterative,grave2018unsupervised}.
  \label{tab:nn}}
\end{table*}

\section{Experiments}
\label{sec:results}

This section reports the main results obtained with our method.
We provide complementary results and an ablation study in the appendix.
We refer to our method without constraints as RCSLS and as RCSLS+spectral if the spectral constraints are used.

\subsection{Implementation details}
We choose a learning rate in $\{1,~10,~25,~50\}$ and a number of epochs in $\{10,~20\}$ on the validation set.
For the unconstrained RCSLS, a small $\ell_2$ regularization can be added to prevent the norm of $\mathbf{W}$ to diverge.
In practice, we do not use any regularization.
For the English-Chinese pairs (en-zh), we center the word vectors.
The number of nearest neighbors in the CSLS loss is $10$.
We use the $\ell_2$-normalized fastText word vectors by Bojanowski \etal \shortcite{bojanowski2017enriching} trained on Wikipedia.

\subsection{The MUSE benchmark} 
Table~\ref{tab:nn} reports the comparison of RCSLS with standard supervised and unsupervised approaches on 5 language pairs (in both directions) of the MUSE benchmark~\cite{conneau2017word}.
Every approach uses the Wikipedia fastText vectors and supervision comes in the form of a lexicon composed of $5$k words and their translations.
Regardless of the relaxation, RCSLS outperforms the state of the art by, on average, $3$ to $4\%$ in accuracy.
This shows the importance of using the same criterion during training and inference.
Note that the refinement step (``refine'') also uses CSLS to finetune the alignments but leads to a marginal gain for supervised methods.

Interestingly, RCSLS achieves a better performance without constraints ($+0.8\%$) for all pairs.
Contrary to observations made in previous works, this result suggests that preserving the distance between word vectors is not essential for word translation.
Indeed, previous works used a $\ell_2$ loss where, indeed, orthogonal constraints lead to an improvement of $+5.3\%$ (Procrustes versus Least Square Error).
This suggests that a linear mapping $\mathbf{W}$ with no constraints works well only if it is learned with a proper criterion.

\begin{table}[h!]
\centering
\begin{tabular}{l c c c c c }
\toprule
  &en-es  &en-fr  &en-de  &en-ru  & avg.\\
\midrule
  Train & 80.7 & 82.3 & 74.8 & 51.9 & 72.4\\
  Ext.  &\textbf{84.1} &  \textbf{83.3} &  \textbf{79.1}  & \textbf{57.9} & \textbf{76.1}\\
\bottomrule
\end{tabular}
\caption{
  Accuracy with and without an extended normalization for RCSLS.
  ``Ext.'' uses the full $200$k vocabulary and ``Train'' only uses the pairs from the training lexicon.
\label{tab:norm}}
\end{table}

\paragraph{Impact of extended normalization.}
Table~\ref{tab:norm} reports the gain brought by including words not in the lexicon (unannotated words) to the performance of RCSLS.
Extending the dictionary significantly improves the performance on all language pairs.

\begin{table}[h]
\centering
\begin{tabular}{l cc }
\toprule
  & en-it & it-en\\
\midrule
Adversarial + refine + CSLS & 45.1 & 38.3 \\
\midrule
\citet{mikolov2013exploiting} & 33.8 & 24.9\\
\citet{dinu2014improving} & 38.5 & 24.6\\
\citet{artetxe2016learning} & 39.7 & 33.8\\
\citet{smith2017offline} & 43.1 & 38.0\\
Procrustes + CSLS & 44.9 & \textbf{38.5}\\
\midrule
RCSLS & \textbf{45.5} & 38.0\\
\bottomrule
\end{tabular}
  \caption{
    Accuracy on English and Italian with the setting of~\citet{dinu2014improving}.
    ``Adversarial'' is an unsupervised technique.
    The adversarial and Procrustes results are from~\citet{conneau2017word}.
  We use a CSLS criterion for retrieval.
  }\label{tab:dinu}
\end{table}

\subsection{The WaCky dataset}
\citet{dinu2014improving} introduce a setting where word vectors are learned on the WaCky datasets~\citep{baroni2009wacky} and aligned with a noisy bilingual lexicon.
We select the number of epochs within $\{1,2,5,10\}$ on a validation set.
Table~\ref{tab:dinu} shows that RCSLS is on par with the state of the art. RCSLS is thus robust to relatively poor word vectors and noisy lexicons.

\begin{table}[h]
\setlength{\tabcolsep}{4.4pt} 
\centering
\begin{tabular}{l ccc c ccc}
\toprule
  & \multicolumn{3}{c}{Original} && \multicolumn{3}{c}{Aligned} \\
\cmidrule{2-4}
\cmidrule{6-8}
   & Sem. & Synt. & Tot. && Sem. & Synt. & Tot.\\
\midrule
  \textsc{Cs} & 26.4 & 76.7 & 63.7 && \textbf{27.3} & \textbf{77.7} & \textbf{64.6}\\
  \textsc{De} & \textbf{62.2} & 56.9 & \textbf{59.5} && 61.4 & \textbf{57.1} & 59.3\\
  \textsc{Es} & 54.5 & 59.4 & 56.8 && \textbf{55.1} & \textbf{61.1} & \textbf{57.9}\\
  \textsc{Fr} & \textbf{76.0} & 54.7 & \textbf{68.5} && 75.2 & \textbf{55.1} & 68.1\\
  \textsc{It} & 51.8 & 62.0 & 56.9 && \textbf{52.7} & \textbf{63.8} & \textbf{58.2}\\
  \textsc{Pl} & 49.7 & 62.4 & 53.4 && \textbf{50.9} & \textbf{63.2} & \textbf{54.5}\\
  \textsc{Zh} & 42.6 & -    & 42.6 && \textbf{47.2} & -    & \textbf{47.2}\\
\midrule
  Avg. & 51.9 & 62.0 & 57.3 && \textbf{52.8} & \textbf{58.5} & \textbf{58.5}\\
\bottomrule
\end{tabular}
  \caption{
    Performance on word analogies for different languages.
    We compare the original embeddings to their mapping to English.
    The mappings are learned using the full MUSE bilingual lexicons.
    We use the fastText vectors of \citet{bojanowski2017enriching}.
    }
  \label{tab:othersym}
\end{table}

\begin{table}[h!]
\centering
\begin{tabular}{l c c c c c}
\toprule
  & BP$^*$ & MUSE & Proc.& \multicolumn{2}{c}{RCSLS} \\
\midrule
  && Orig. & Full & Orig. & Full \\
  \cmidrule{5-6}
  \textsc{Bg} & 55.7 & 57.5 & 58.1 & 63.9 & \textbf{65.2}\\
  \textsc{Ca} & 66.5 & 70.9 & 70.5 & 73.8 & \textbf{75.0}\\
  \textsc{Cs} & 63.9 & 64.5 & 66.3 & 68.2 & \textbf{71.1}\\
  \textsc{Da} & 66.8 & 67.4 & 68.3 & 71.1 & \textbf{72.9}\\
  \textsc{De} & 68.9 & 72.7 & 73.5 & 76.9 & \textbf{77.6}\\
  \textsc{El} & 54.9 & 58.5 & 60.1 & 62.7 & \textbf{64.5}\\
  \textsc{Es} & 82.1 & 83.5 & 84.5 & 86.4 & \textbf{87.1}\\
  \textsc{Et} & 41.5 & 45.7 & 47.3 & 49.5 & \textbf{53.7}\\
  \textsc{Fi} & 56.7 & 59.5 & 61.9 & 65.8 & \textbf{69.9}\\
  \textsc{Fr} & 81.7 & 82.4 & 82.5 & \textbf{84.7} & \textbf{84.7}\\
  \textsc{He} & 51.5 & 54.1 & 55.4 & 57.8 & \textbf{60.0}\\
  \textsc{Hr} & 48.9 & 52.2 & 53.4 & 55.6 & \textbf{60.2}\\
  \textsc{Hu} & 61.9 & 64.9 & 66.1 & 69.3 & \textbf{73.1}\\
  \textsc{Id} & 62.8 & 67.9 & 67.9 & 69.7 & \textbf{72.9}\\
  \textsc{It} & 75.3 & 77.9 & 78.5 & 81.5 & \textbf{82.8}\\
  \textsc{Mk} & 53.9 & 54.6 & 55.4 & 59.9 & \textbf{60.4}\\
  \textsc{Nl} & 72.0 & 75.3 & 76.1 & 79.7 & \textbf{80.5}\\
  \textsc{No} & 65.3 & 67.4 & 68.3 & 71.2 & \textbf{73.3}\\
  \textsc{Pl} & 63.3 & 66.9 & 68.1 & 70.5 & \textbf{73.5}\\
  \textsc{Pt} & 77.7 & 80.3 & 80.4 & 82.9 & \textbf{84.6}\\
  \textsc{Ro} & 66.3 & 68.1 & 67.6 & \textbf{74.0} & 73.9\\
  \textsc{Ru} & 61.3 & 63.7 & 64.3 & 67.1 & \textbf{70.3}\\
  \textsc{Sk} & 55.1 & 55.3 & 57.9 & 59.0 & \textbf{61.7}\\
  \textsc{Sl} & 51.1 & 50.4 & 52.5 & 54.2 & \textbf{58.2}\\
  \textsc{Sv} & 55.9 & 60.0 & 64.0 & 63.7 & \textbf{69.5}\\
  \textsc{Tr} & 57.4 & 59.2 & 61.4 & 61.9 & \textbf{65.8}\\
  \textsc{Uk} & 48.7 & 49.3 & 51.3 & 51.5 & \textbf{55.5}\\
  \textsc{Vi} & 35.0 & 55.8 & 63.0 & 55.8 & \textbf{66.9}\\
\midrule
CSLS & 60.8 & 63.8 & 65.2 & 67.4 & \textbf{70.2}\\
\midrule
NN   & 54.6 & 57.4 & 57.5 & 62.4 & \textbf{68.5}\\
\bottomrule
\end{tabular}
  \caption{Comparison with publicly available aligned vectors over 28 languages.
    All use supervision.
    Alignements are learned either on the ``Original'' or ``Full'' MUSE training.
    We report the detailed performance with a CSLS criterion and the average for both NN and CSLS criteria. \newline
  $^*$BP uses a different training set of comparable size.
  }\label{tab:MUSE}
\end{table}

\subsection{Comparison with existing aligned vectors}
Recently, word vectors based on fastText have been aligned and released by Smith \etal~\shortcite[BabylonPartners, BP]{smith2017offline}~
and Conneau \etal~\shortcite[MUSE]{conneau2017word}.
Both use a variation of Procrustes to align word vectors in the same space.

We compare these methods to RCSLS and report results in Table~\ref{tab:MUSE}.
RCSLS improves the performance by $+3.5\%$ over MUSE vectors when trained with the same lexicon (Original).
Training RSCSL on the full training lexicon (Full) brings an additional improvement of $+2.9\%$ on average with a CSLS criterion, and $+6.1\%$ with a NN criterion.
For reference, the performance of Procrustes only improves by $+1.4\%$ with CSLS and even degrades with a NN criterion.
RCSLS benefits more from additional supervision than Procrustes.
Finally, the gap between RCSLS and the other methods is higher with a NN criterion, suggesting that RCSLS imports
some of the properties of CSLS to the dot product between aligned vectors.


\begin{table}[h!]
\centering
\begin{tabular}{l r cc}
\toprule
&&	Original	& Aligned\\
\midrule
  \multirow{3}*{\textsc{De}}  & \textsc{Gur350}& 72& \textbf{74}\\
  & \textsc{Ws350 }& 68& \textbf{70}\\
  & \textsc{Zg222 }& \textbf{46}& 44\\
\midrule
  \textsc{Es} & \textsc{Ws353} & 59	& \textbf{61}\\
\midrule
  \multirow{1}*{\textsc{It}} & \textsc{Ws350} & \textbf{64}& \textbf{64}\\
\midrule
  \textsc{Pt} & \textsc{Ws353} & \textbf{60} & 58\\
\midrule
  \textsc{Ro} & \textsc{Ws353} & \textbf{58} & 55\\
\midrule
  &\textsc{Hj} & \textbf{67} & 65\\
  &\textsc{Ws350} & \textbf{59} & 58\\
\midrule
  \textsc{Zh} & \textsc{Sim}& 35	& \textbf{44}\\
\midrule
  Avg. && \hspace{7pt}58.8 & \hspace{7pt}\textbf{59.2}\\
\bottomrule
\end{tabular}
  \caption{
    Performance on word similarities for different languages.
    We compare the original embeddings to their mapping to English.
    The mappings are learned with the full MUSE bilingual lexicons over the fastText vectors of \citet{bojanowski2017enriching}.
    }
  \label{tab:similarities}
\end{table}

\subsection{Impact on word vectors}
Non-orthogonal mapping of word vectors changes their dot products.
We evaluate the impact of this mapping on word analogy tasks~\cite{mikolov2013efficient}.
In Table~\ref{tab:othersym}, we report the accuracy on analogies for raw word vectors and our vectors mapped to English with an alignement trained on the full MUSE training set.
Regardless of the source language, the mapping does not negatively impact the word vectors.
Similarly, our alignement has also little impact on word similarity, as shown in Table~\ref{tab:similarities}.

We confirm this observation by running the reverse mapping, i.e., by mapping the English word vectors of~\citet{mikolov2018advances} to Spanish. It leads to an improvement of $1\%$ both for vectors trained on Common Crawl ($85\%$ to $86\%$) and Wikipedia + News ($87\%$ to $88\%$).


\section{Conclusion}
This paper shows that minimizing a convex relaxation of the CSLS loss significantly improves the quality of bilingual word vector alignment.
We use a reformulation of CSLS
that generalizes to convex functions beyond dot-products and provides to a single end-to-end training that is consistent with the inference stage.
Finally, we show that removing the orthogonality constraint does not degrade the quality of the aligned vectors.

\paragraph{Acknowledgement.} We thank Guillaume Lample and Alexis Conneau for their feedback and help with MUSE.

\clearpage
\bibliography{example_paper}
\bibliographystyle{acl_natbib}

\clearpage
\begin{appendices}

\section{Ablation study}

This appendix presents an ablation study, to validate the design choices that we made.

\begin{figure}[h]
  \centering
  \includegraphics[width=.94\linewidth]{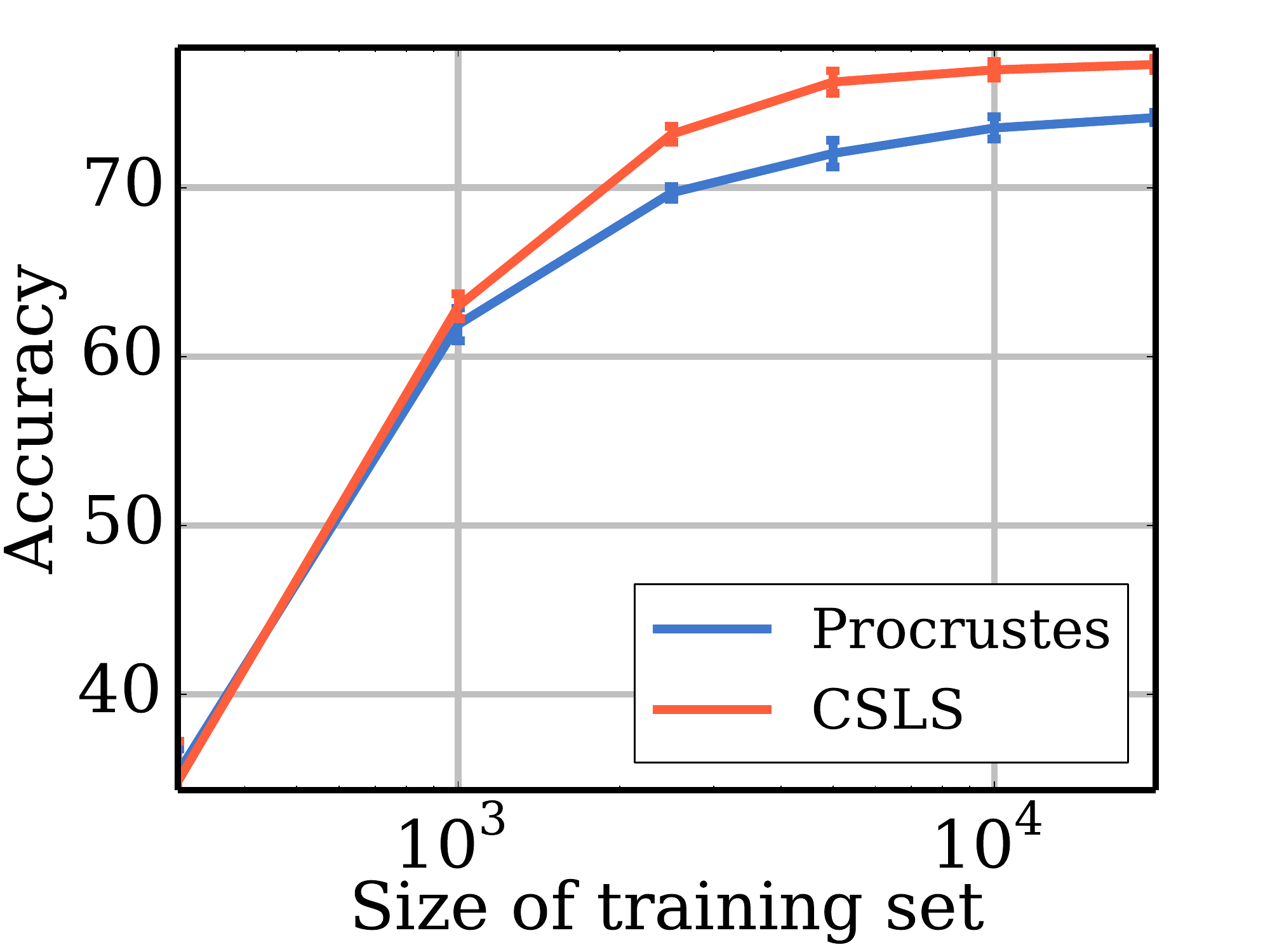}
  \caption{Accuracy as a function of the training set size (log scale) on the en-de pair.}\label{fig:ende}
\end{figure}

\paragraph{Size of training lexicon.}
Figure~\ref{fig:ende} compares the accuracy of RCSLS and Procrustes as a function of the training set size.
On small training sets, the difference between RCSLS and Procrustes is marginal but increases with the training set size.

\begin{figure}[h]
  \centering
  \includegraphics[width=.94\linewidth]{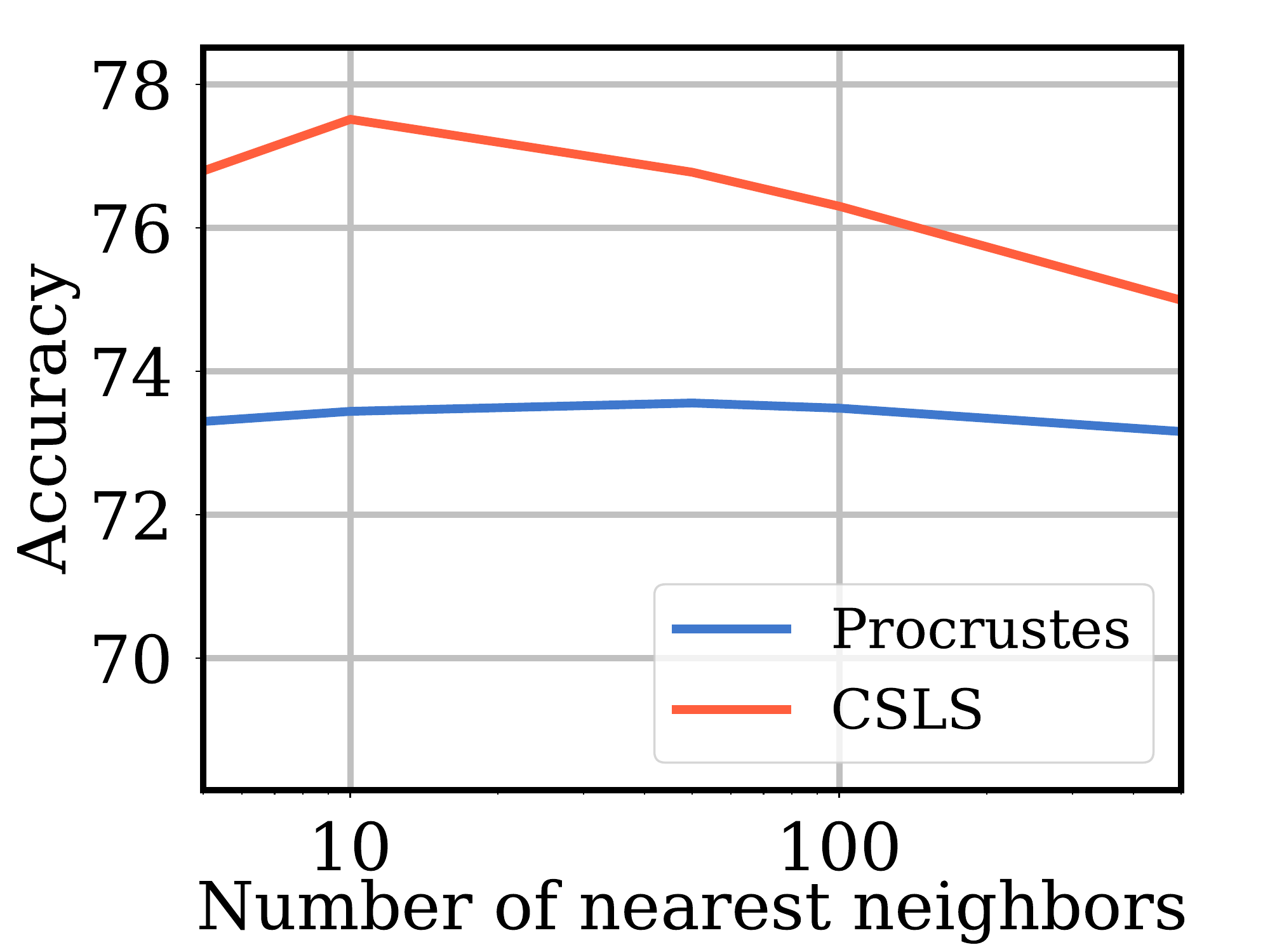}
  \caption{Accuracy as a function of the number of nearest neighbors, averaged over $8$ different pairs.}\label{fig:knn}
\end{figure}

\paragraph{Impact of the number of nearest neighbors.}
The CSLS criterion and the RCSLS loss are sensible to the number of nearest neighbors.
Figure~\ref{fig:knn} shows the impact of this parameter on both Procrustes and RCSLS.
Procrustes is impacted through the retrieval criterion while RCSLS is impacted by the loss and the criterion.
Taking $10$ nearest neighbors is optimal and the performance decreases significantly with a large number of neighbors.

\begin{table}[h!]
\centering
\begin{tabular}{l c c c c}
\toprule
  & en-es & es-en & en-ru & ru-en\\
\midrule
Linear    & 84.1 & 86.3 & 58.0 & 67.2 \\
logSumExp & 84.1 & 86.3 & 58.3 & 67.0 \\
\bottomrule
\end{tabular}
  \caption{Comparison between different functions in CSLS on four language pairs. Linear is the standard criterion, while logSumExp is equivalent to a logistic regression with hard mining.}\label{tab:log}
\end{table}

\paragraph{Comparison of alternative criterions.}
As discussed in the main paper, the dot product in the CSLS terms can be replaced by any convex function of $\mathbf{W}$ and still yield a convex objective.
Using a logSumExp function, i.e., $f(x) = \log(\sum_i(\exp(x_i)))$ is equivalent to a ``local'' logistic regression classifier, or equivalently, to a logistic regression with hard mining.
In this experiment, we train our model using the alternative loss and report the accuracy of the resulting lexicon in Table~\ref{tab:log}.
We observe that this choice does not significantly changes the performance. This suggests that the local property of the criterion is more important than the form of the loss.

\begin{table}[h!]
\centering
\setlength{\tabcolsep}{4.4pt} 
\begin{tabular}{l c c c c c}
\toprule
&& \multicolumn{2}{c}{Original} & \multicolumn{2}{c}{Full} \\
\cmidrule(r){3-4} \cmidrule(r){5-6}
  & BP & MUSE & RCSLS &  Proc.& RCSLS \\
\midrule
  \multicolumn{6}{l}{\emph{with exact string matches}}\\
  NN   & 54.6 & 57.4 & 62.4 & 57.5 & \textbf{68.5}\\
  CSLS & 60.8 & 63.8 & 67.4 & 65.2 & \textbf{70.2}\\
\midrule
  \multicolumn{6}{l}{\emph{without exact string matches}}\\
  NN   & 56.6 & 55.5 & 61.4 & 53.7 & \textbf{64.3} \\
  CSLS & 61.5 & 60.4 & 65.4 & 60.2 & \textbf{65.7} \\
\bottomrule
\end{tabular}
  \caption{Comparison with publicly available aligned vectors, averaged over $28$ language pairs. All use supervision.
    Alignements are learned either on the ``Original'' or ``Full'' MUSE training.
    We report performance with the NN and CSLS criterion on either the full MUSE test set or without the exact string matches.
    BP uses a different training set with $5$k words.}
  \label{tab:MUSE}
\end{table}

\begin{table*}[t]
\setlength{\tabcolsep}{4.8pt} 
\centering
\begin{tabular}{l c cc cc cc cc  cc c}
\toprule
Method              &en-es&es-en&en-fr&fr-en&en-de&de-en&en-ru&ru-en&en-zh&zh-en& avg.\\
\midrule
Adv. + ref. + NN& 79.1 & 78.1 & 78.1 & 78.2 &71.3 & 69.6 &  37.3 & 54.3 & 30.9 & 21.9 & 59.9\\
Adv. + ref. + CSLS & 81.7& 83.3& 82.3& 82.1& 74.0& 72.2& 44.0& 59.1& 32.5 & 31.4 & 64.3\\
\midrule
Procrustes + NN  & 77.4 &  77.3 &74.9 & 76.1 &  68.4 & 67.7 &  47.0 & 58.2 & 40.6 & 30.2& 61.8\\
Procrustes + CSLS& 81.4 & 82.9 & 81.1 & 82.4 & 73.5 & 72.4 & 51.7 & 63.7 &42.7 & 36.7& 66.8\\
\midrule
RCSLS + NN& 81.1 & 84.9 &  80.5 & 80.5 &  75.0 & 72.3  & 55.3  &  67.1 & 43.6  & 40.1 & 68.0\\
RCSLS + CSLS& 84.1 & 86.3 &  83.3 & 84.1 &  79.1 & 76.3  & 57.9  &  67.2 & 45.9  & 46.4 & 71.0\\
\bottomrule
\end{tabular}
\caption{
  Comparison between a nearest neighbor (NN) criterion and CSLS.
  \label{tab:nn}}
\end{table*}

\paragraph{Exact string matches.}
The MUSE datasets contains exact string matches based on vocabularies built on Wikipedia. The matches may reflects correct
translations but can come from other sources, like English word that frequently appears in movie or song titles.
Table~\ref{tab:MUSE} compares alignments on the MUSE test set with and without exact string matches average over $28$ languages.
Note that we do not remove exact matches in the training sets for fair comparison with MUSE vectors.
We note that the gap between our vectors and others is more important with an NN criterion.
We also observe that, the performance of all the methods drop when the exact string matches are removed.

\paragraph{Impact of the retrieval criterion.} Table~\ref{tab:nn} shows performance on MUSE with a nearest neighbors (NN) criterion.
  Replacing CSLS by NN leads to a smaller drops for RCSLS ($3\%$) than for competitors (around $5\%$),
  suggesting that RCSLS transfers some local information encoded in the CSLS criterion to the dot product.

\section{Alignment and word vectors}
In this appendix, we look at the relation between the quality of the word vectors and the quality of an alignment.
We measure both the impact of the vectors on the alignment and the impact of a non-orthogonal mapping on word vectors.

\begin{table}[h]
\centering
\begin{tabular}{l c c}
\toprule
& without subword & with subword \\
\midrule
en-es & 82.8 & 84.1 \\
es-en & 84.1 & 86.3 \\
en-fr & 82.3 & 83.3 \\
fr-en & 82.5 & 84.1 \\
en-de & 78.5 & 79.1 \\
de-en & 74.1 & 76.3 \\
\bottomrule
\end{tabular}
  \caption{Impact on the alignment of the quality of the word vectors trained on the same corpora.}\label{tab:qual}
\end{table}

\paragraph{Quality of the embedding model.}
In this experiment, we study the impact of the quality of the word vectors on the performance of word translation.
For this purpose, we trained word vectors on the same Wikipedia data, using skipgram with and without subword information.
In Table~\ref{tab:qual}, we report the accuracy for different language pairs when using these two sets of word vectors.
Overall, we observe that using subword information improves the accuracy by a few points on all pairs.

\begin{table}[h]
\centering
\begin{tabular}{l c c c}
\toprule
  & Sem. & Synt. & Tot. \\
\midrule
Orig. & 79.4 & 73.4 & 76.1 \\
\midrule
en$\rightarrow$es & \textbf{80.5} & 75.8 & \textbf{78.0} \\
en$\rightarrow$fr & 79.8 &\textbf{75.9} & 77.6 \\
en$\rightarrow$de & 80.0 &\textbf{75.9} & 77.6 \\
en$\rightarrow$ru & 79.5 &74.6 & 76.8 \\
\bottomrule
\end{tabular}
\caption{Semantic and syntactic accuracies of English vectors and their mappings to different languages.
  \label{tab:sym}}
\end{table}

\paragraph{Impact on English word vectors.}
We evaluate the impact of a non-orthogonal mapping on the English word analogy task~\cite{mikolov2013efficient}.
Table~\ref{tab:sym} compares on analogies the raw English word vectors to their alignments to $4$ languages.
Regardless of the target language, the mapping does not have negative impact on the word vectors.

\begin{table}[h]
\centering
\begin{tabular}{l c c c}
\toprule
  & en-es. & en-de & en-it \\
\midrule
  NASARI baseline & 0.64 & 0.60 & 0.65\\
  BP & 0.72 & 0.69 & 0.71\\
  MUSE            & 0.71 & 0.71 & 0.71\\
  \midrule
  RCSLS            & 0.71 & 0.71 & 0.71\\
\bottomrule
\end{tabular}
  \caption{
    Cross-lingual word similarity on the NASARI datasets of~\citet{camacho2016nasari}. We report the Pearson correlation.
  BP, MUSE and RCSLS uses the Wikipedia fastText vectors.}\label{tab:cross}
\end{table}

\paragraph{Cross-lingual similarity.} Finally, we evaluate our aligned vectors on the task of cross-lingual word similarity in Table~\ref{tab:cross}.
They obtain similar results to vectors aligned with an orthogonal matrix.
These experiments concur with the previous observation that a linear non-orthogonal mapping does not hurt the geometry of the word vector space.

\end{appendices}

\end{document}